\documentclass[review]{fcs}

\usepackage{graphicx}
\usepackage{booktabs}

\usepackage{bm}
\usepackage{algorithm}  
\usepackage{multicol}
\usepackage{multirow}
\usepackage{mdframed}
\usepackage{pgfplots}
\usepackage{tcolorbox}
\usepackage{pgf-pie}  
\usepackage{subcaption} 
\usepackage{wrapfig}
\usepackage[noend]{algpseudocode}

\usepackage{makecell}
\tcbuselibrary{skins,breakable} 

\usepackage{tabularx}
\usepackage{array}
\newtcolorbox{myshadowbox}{  
    enhanced,  
    breakable,
    drop shadow southeast,  
    sharp corners,  
    colback=white,  
    colframe=black  
}  
\newcolumntype{Y}{>{\centering\arraybackslash}X}

\definecolor{BlueViolet}{rgb}{0.21,0.49,0.74}

\title{CCA: Collaborative Competitive Agents for Image Editing}
\author[1]{Tiankai Hang}
\author[2]{Shuyang Gu}
\author[2]{Dong Chen}
\author*[1]{Xin Geng}
\author*[1,2]{Baining Guo}
\address[1]{School of Computer Science and Engineering, Southeast University, Nanjing 211189, China}
\address[2]{Microsoft Research Asia}
\fcssetup{
  received       = {month dd, yyyy},
  accepted       = {month dd, yyyy},
  corr-email     = {xgeng@seu.edu.cn, 307000167@seu.edu.cn},
}
\begin{abstract}
  This paper presents a novel generative model, Collaborative Competitive Agents (CCA), which leverages the capabilities of multiple Large Language Models (LLMs) based agents to execute complex tasks. Drawing inspiration from Generative Adversarial Networks (GANs), the CCA system employs two equal-status generator agents and a discriminator agent. The generators independently process user instructions and generate results, while the discriminator evaluates the outputs, and provides feedback for the generator agents to further reflect and improve the generation results. Unlike the previous generative model, our system can obtain the intermediate steps of generation. This allows each generator agent to learn from other successful executions due to its transparency, enabling a collaborative competition that enhances the quality and robustness of the system's results. The primary focus of this study is image editing, demonstrating the CCA's ability to handle intricate instructions robustly. The paper's main contributions include the introduction of a multi-agent-based generative model with controllable intermediate steps and iterative optimization, a detailed examination of agent relationships, and comprehensive experiments on image editing.
\end{abstract}

\begin{document}
\sloppy 
\section{Introduction}

The human endeavor to conceptualize Artificial Intelligence (AI) is fundamentally rooted in the aspiration to engineer intelligent entities. In the contemporary era, this pursuit has been significantly propelled by the evolution and advancement of Large Language Models (LLMs)~\cite{gpt4v,openai2023gpt4,ouyang2022instructgpt,touvron2023llama,touvron2023llama2}. The rapidly developing LLM-based agents~\cite{yao2022webshop,chatdev,swan2023math} have outpaced their predecessors, demonstrating a higher degree of intelligence through a more sophisticated understanding of human intentions and a greater competency in assisting with complex tasks. This progression has instigated a technological revolution across a multitude of fields, encompassing software development~\cite{chatdev}, education~\cite{kalvakurthi2023hey,swan2023math}, sociology~\cite{park2023generative}, among others.

When examining the realm of generative models, specifically Generative Adversarial Networks (GANs)~\cite{goodfellow2014generative,brock2018biggan,karras2019stylegan,Karras2019stylegan2} and diffusion models~\cite{ho2020ddpm,song2021scoresde,dhariwal2021adm,karras2022edm,podell2023sdxl,nichol2021iddpm,hang2024noise-schedule,wangfine}, we encounter two notable challenges. The first challenge is the models' limited ability to process complex, compound tasks. To illustrate, consider a task that involves ``colorizing an old photograph, replacing the depicted individual with the user's image, and adding a hoe in the user's hand''. Such a multifaceted task surpasses the capability of even the most advanced generative models. The second challenge arises in the update process of a generated result. This process is contingent upon the preservation of the compute graph. However, the sheer volume of results generated by diverse algorithms makes maintaining this compute graph a significant hurdle. Consequently, this creates a barrier to learning from other generative models, given their black-box nature.

In this paper, we introduce a novel generative model that harnesses the capabilities of multiple LLM-based agents, which effectively circumvents these two challenges. Leveraging the agents' powerful task decomposition abilities, our model can efficiently manage highly complex tasks. Simultaneously, during the generation process, we can extract insights into how the agents comprehend, dissect, and execute the task, enabling us to modify internal steps and enhance the results. Crucially, the model's transparency allows the agent to learn from successful executions by other agents, moving away from the black-box model paradigm. We underscore that this transparency is a pivotal factor contributing to the enhanced quality and robustness of the system. 

Generative Adversarial Networks (GANs)~\cite{goodfellow2014generative} can be viewed as an early endeavor to incorporate a multi-agent system into generative models. GANs utilize two agents, namely, a generator and a discriminator. A cleverly designed optimization function allows these agents to learn from their adversarial interaction, ideally reaching a Nash equilibrium. Similarly, in our multi-agent system, we have discovered that the establishment of relationships between different agents is a critical determinant of success.

Drawing inspiration from GANs, our system employs two generators and one discriminator. The two generator agents, of equal status, independently process user instructions and generate results. The discriminator agent then evaluates these generated results, providing feedback to each generator and determining which result is superior. The generator agents have dual responsibilities. Firstly, they must reflect on the feedback from the discriminator. Secondly, they should consider the results produced by the other generator agent to enhance their generation process. This process is iteratively carried out until the discriminator deems the best result to have sufficiently met the user's requirements. We underscore that through this collaborative competition, the two generators can continuously augment the quality and robustness of the system's results. Consequently, we have named our system Collaborative Competitive Agents (CCA).

In this paper, we concentrate on image editing, although our Collaborative Competitive Agents (CCA) system is a versatile generative model. Conventional image editing methods~\cite{brooks2023instructpix2pix,hertz2022prompt-to-prompt,meng2021sdedit} fall short when dealing with intricate instructions, resulting in less robust outcomes. Our proposed generative model can considerably enhance this situation through the collaborative competition of multiple agents.

In summary, our primary contributions are as follows:
\begin{enumerate}
    \item We introduce a new generative model based on multiple agents, which features controllable intermediate steps and can be iteratively optimized.
    \item We have meticulously examined the relationships among multiple agents, highlighting that reflection, cooperation, and competition are integral to the system's quality and robustness.
    \item We have conducted comprehensive experiments on image editing, demonstrating for the first time the ability to robustly handle complex instructions.
\end{enumerate}

\section{Related Work}
\subsection{Large Language Model-based Agents}

Agents are artificial entities capable of perceiving the environment, making decisions, and taking actions to accomplish specific goals~\cite{sutton2018reinforcement,agent-survey,weng2023prompt,deng2024composerx}.
Recent advancements in Large Language Models (LLMs) have demonstrated significant intelligence~\cite{schick2023toolformer,YongliangWU:in-context}, offering promising avenues for the evolution of intelligent agents.
LLM-based agents possess the ability to memorize, plan, and utilize tools. The ``memory'' feature allows these agents to store sequences of past observations, thoughts, and actions for future retrieval. The CoT~\cite{wei2022chainofthoughts} enhances the LLM's capacity to solve complex tasks by ``thinking step by step''. Moreover, agents employ a reflection mechanism~\cite{yao2022react,madaan2023self,yang2023idea2img} to enhance their planning abilities. Furthermore, LLM-based agents can leverage various tools to interact with their environment~\cite{schick2023toolformer,shen2023hugginggpt,openai2023gpt4}, such as shopping~\cite{yao2022webshop} and coding~\cite{chatdev}. Some studies~\cite{driess2023palm} equip these agents with embodied actions to facilitate interaction with humans.

Similar to human society, a single skilled agent can handle specific tasks, while a multi-agent system can tackle more complex ones. To foster autonomous cooperation among agents, CAMEL~\cite{li2023camel} introduces a novel communicative agent framework called ``role-playing", which incorporates inception prompting. AgentVerse~\cite{chen2023agentverse} introduces a multi-task-tested framework for group agents collaboration, designed to assemble a team of agents that can dynamically adjust to the intricacy of the task at hand. ChatDev~\cite{chatdev} demonstrates significant potential in software development by integrating and unifying key processes through natural language communication. Concurrently, ChatEval~\cite{chan2023chateval} employs multiple agents as a referee team, they engage in debates among themselves and ultimately determine a measure of the quality of the LLM generation. 
Pure collaboration means that agents work together, completing their respective parts to achieve a common goal. Competition, on the other hand, implies rivalry, where each agent's objective is to pursue their own success, making their own plans and decisions based on feedback from both themselves and other agents. Therefore, this is not a situation of complete cooperation. We refer to this type of competition as collaborative competition.
In this paper, we explore a scenario where multiple agents with collaborative competition to achieve goals.

\subsection{Image Editing}

Image Editing has been a long-standing vision task and is widely used in real-world applications~\cite{brooks2023instructpix2pix,Geng23instructdiff,hang2023language,hertz2022prompt-to-prompt,mokady2023nulltext}. The primary objective of this task is to manipulate an image to align with user-specified requirements. 
Traditional methods mainly address specific tasks like style transfer~\cite{johnson2016perceptual,gatys2016styletransfer,gu2018arbitrary,ding2024regional}, image translation~\cite{zhu2017cyclegan,isola2017image}, and object removal or replacement~\cite{bertalmio2000imageinpainting,criminisi2003object,sun2005imagecompletion,yang2023paint}. 
Later works~\cite{Zhang2023MagicBrush, Geng23instructdiff} utilize text-to-image models to perform edits from user-provided text instructions.

From the generative model perspective, many early studies leveraged the outstanding disentangled properties in the latent space of GANs~\cite{karras2019stylegan, Karras2019stylegan2}, altering image attributes via latent code manipulation~\cite{xia2022gan-inversion-survey,shen2020interpreting,zhu2020indomain}. 
Some studies~\cite{patashnik2021styleclip} use CLIP~\cite{radford2021clip} to facilitate text-driven image editing.
Recent advancement in diffusion models~\cite{ho2020ddpm,dhariwal2021adm,karras2022edm,song2021scoresde,gu2022vector} has demonstrated great success in image generation \cite{rombach2022ldm,ramesh2022dalle2,saharia2022imagen,balaji2022ediffi}, with these pre-trained diffusion models widely used in image editing. Meng \textit{et al.}~\cite{meng2021sdedit} proposed reversing and perturbing the SDE to conduct image manipulation, while EDCIT~\cite{wallace2023edict} enhanced editing quality through more precise inversion. Prompt-to-Prompt~\cite{hertz2022prompt-to-prompt} investigated the role of words in attention and edited images by modifying the cross-attention. Null-text Inversion~\cite{mokady2023nulltext} progresses by optimizing tokens to avoid artifacts from classifier-free guidance. Dreambooth~\cite{ruiz2023dreambooth} fine-tuned the pre-trained text-to-image diffusion model to perform subject-driven generation. 
More recent work, VisProg~\cite{gupta2023visual}, leverages the \textit{in-context learning capabilities} of LLMs to perform image editing. However, it\textit{ heavily relies on the given examples} and choices for the most suitable tool. In contrast to it, our proposed CCA is a multi-agent system that not only includes tools for planning and execution but also iteratively improves the outcomes based on feedback through collaboration and competition.

\section{Method}

\begin{figure*}[!t]
    \centering
    \includegraphics[width=1\linewidth]{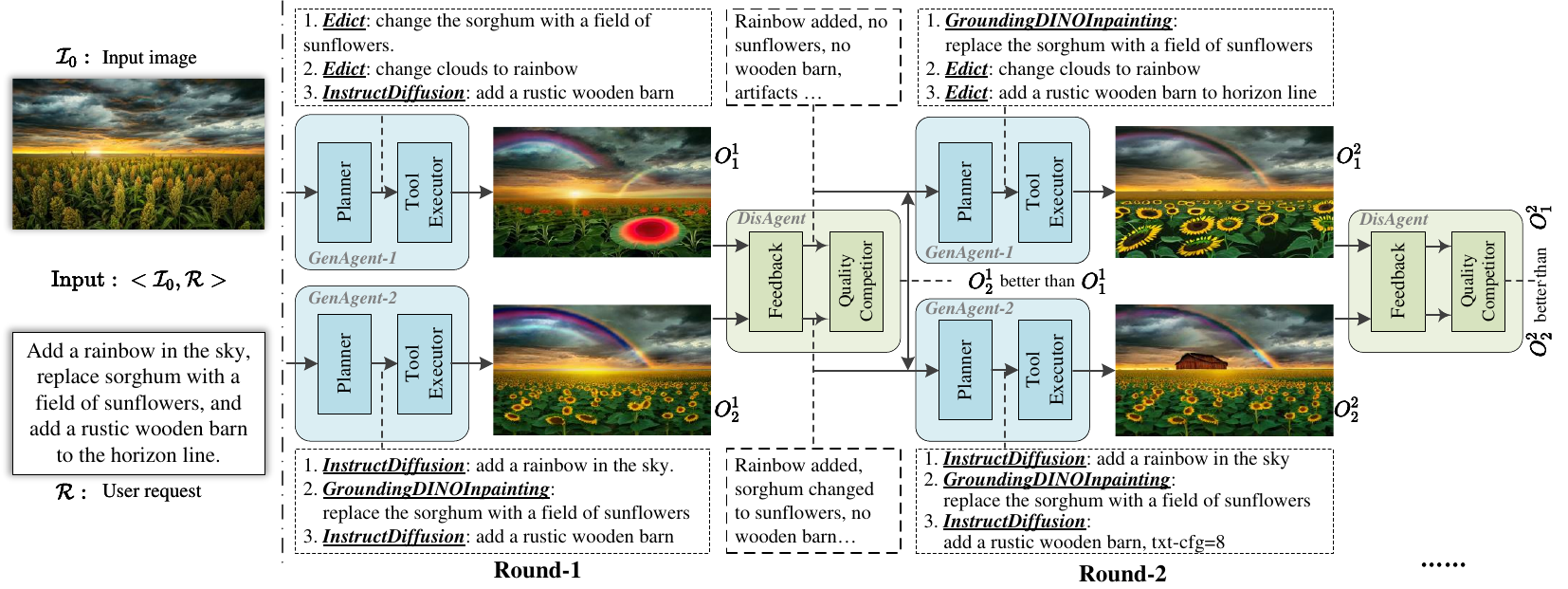}
    \caption{The framework of our Collaborative Competitive Agents system. Through providing feedback, the discriminator agent encourages the generator agent to engage in both collaborative learning and competition. The system's performance undergoes iterative optimization to effectively meet user requirements.}
    \label{fig:framework}
\end{figure*}

Our framework, identified as CCA, incorporates two distinct types of agents: the Generator Agent and the Discriminator Agent. 

\subsection{Generator Agent}\label{sec:generator-agent}

The Generator Agent edits the image through the utilization of two core modules: the \textbf{Planner}, which is engaged in deciphering the user's request and making plans, and the \textbf{Tool Executor}, responsible for systematically modifying the image in a step-by-step manner. 

\begin{algorithm}[!h]  
    \caption{Algorithm for CCA Framework}  
    \begin{algorithmic}[1]  
    \State \textbf{Input:} User request $R$, tool set $T$, input image $\mathcal{I}_0$, number of rounds $M$  
    \State \textbf{Output:} Edited image $\mathcal{I}_{\text{edited}}$  
    \State \textbf{Other Symbols:} Memory bank $\mathcal{M}$, Feedback $F$, feedback of each subtask $\{f_j\}$, subtasks of the plan $\{s_j\}$, round index $m$, intermediate edited results $\{o_j\}$, generator output $O$, Tool executor $\mathcal{E}$, Planner $\mathcal{P}$, Quality Competitor $\mathcal{QC}$, color \textcolor{BlueViolet}{blue} denotes corresponding item from another generator agent.  
    \State \Function{Agent}:  
    \State $\{s_j\}^{m}  \leftarrow \mathcal{P} ( \emph{R}, \emph{T},\{s_j\}^{m-1}, F^{m-1},$ \textcolor{BlueViolet}{$F^{m-1}$} $)$ \Comment{get plan}  
    \State $o_1 \leftarrow  \mathcal{I}_0$  
    \ForAll{$s_j$ in $\{s_j\}^{m}$} $o_{j+1} \leftarrow \mathcal{E}(s_j, T(t_i), o_j, f_j)$ \Comment{execute the plan}  
    \EndFor  
    \State $O^{m} \leftarrow o_{n+1}$  
    \State $F^{m} \leftarrow \mathcal{FB}(O^{m}, R)$ \Comment{Get the feedback}  
    \State \Return $O^{m}, F^{m}$  
    \EndFunction  
    \Statex {}  
    \State $\mathcal{M}, F, \{f_j\}^{0}, \{s_j\}^{0}  \leftarrow \emptyset$  
    \For{$m=1$ to $M$}  
      
    \State $O^{m}, F^{m} \leftarrow  Agent1(\emph{R}, \emph{T}, \mathcal{I}_0) $; \textcolor{BlueViolet}{$O^{m}, F^{m}$} $ \leftarrow  Agent2(\emph{R}, \emph{T}, \mathcal{I}_0) $
    \State $\{f_j\},$ \textcolor{BlueViolet}{$\{f_j\}$}$ \leftarrow F^{m},$  \textcolor{BlueViolet}{$F^{m}$}  \Comment{Decompose feedback}  
\State $\mathcal{I}_{\text{best}}, F_{\text{best}} = \mathcal{QC}\left( (O^{m}, F^{m}), (\textcolor{BlueViolet}{O^{m},F^{m}}), \mathcal{M} \right)$ \Comment{Quality compete}  
\State $M \leftarrow \{M, (\mathcal{I}_{\text{best}} , F_{\text{best}})\}$  \Comment{Update the memory}  
\If{$R$ is met by $\mathcal{I}_{\text{best}}$}  
\textbf{break}  
\EndIf  
\EndFor  
\State $\mathcal{I}_{\text{edited}} \leftarrow \mathcal{I}_{\text{best}}$  
\end{algorithmic}  
\label{alg:cca}  
\end{algorithm}  

\subsubsection{Planner}\label{sec:planner}

Existing image editing models often grapple with effectively managing complex user requirements. To address it, we employ a Planner agent, denoted as $\mathcal{P}$, to decompose these requirements into several straightforward and clear subtasks. Typically, user requirements encompass an input image \emph{I} and an associated editing goal, \emph{G}. To enable the Large Language model to comprehend the image, we utilize LLaVA-1.5~\cite{liu2023improvedllava15} to get the image caption $\mathcal{C}$, which serves as a preliminary understanding of the input image. For simplicity, We denote these three elements as user requirements, \emph{R}. 

The Planner agent $\mathcal{P}$ takes the requirements \emph{R} as inputs, decomposing them into a sequence of subtasks, represented as $\{ s_j\}_{j=1}^{n}$.
\begin{align}
    \{ s_1, s_2, \ldots, s_n \} &= \mathcal{P} ( \emph{R}, \emph{T} ),
 \end{align}
Each subtask $s_j$ contains a clear goal to be achieved and a selected tool from Toolset $T$ to accomplish this goal. The toolset outlines which tools are available to the agent and their respective functions, which will be detailed further in section \ref{sec:tool-detail}. 

For instance, if the user request is ``\textit{Make the background a county fair and have the man a cowboy hat, 512pix}''. It can be divided into several sequential steps: 1. Subtask $s_1$ involves loading and resizing the image to a resolution of 512 pixels using the \texttt{Resize} tool; 2. Subtask $s_2$ requires Changing the background to a county fair using the editing tool \texttt{Edict}~\cite{wallace2023edict}; 3. Lastly, subtask $s_3$ requires adding a cowboy hat to the man using the \texttt{InstructDiffusion}~\cite{Geng23instructdiff} tool. 

It's plausible that multiple tools could handle the same sub-task, but each tool may have distinct advantages in different scenarios. Consequently, generating an optimal plan on the first attempt can be challenging. In response to this, we have introduced a strategy that enhances the planning process through multiple rounds of reflection. The \emph{reflection} mechanism is designed to incrementally improve the plan to meet the editing requirements by utilizing feedback. The feedback \emph{F} assesses the success of achieving a sub-goal with the chosen tool and determines whether modifications to the sub-goal are necessary. The feedback is generated by the discriminator agent, which will be further discussed in Sec.~\ref{sec:feedback}.

In summary, beyond its primary function, the Planner agent also serves to reflect upon and enhance plans based on the preceding plan:
\begin{align}
    & S^{m} = \mathcal{P} (\emph{R}, \emph{T}, S^{m-1}, F^{m-1}),
\end{align}
where the superscript $m$ denotes the current round of plan. In the initial round, when $m$ equals 1, there are neither feedbacks nor previous plans available, thus both $S^{(0)}$ and $F^{(0)}$ are set to $\emptyset$ (empty).

\subsubsection{Executor}\label{sec:tool-executor}
When the Planner agent generates a detailed plan that specifies which tool should be employed for each task, we engage another agent, the Executor $\mathcal{E}$, to use the corresponding tools to sequentially execute the plan $\{ s_1, s_2, \ldots, s_n \}$. For each individual subtask $s_j$, the Executor should meticulously explore how to optimally leverage the tool to accomplish it.

For its initial run, the Executor should carefully review the tool's detailed instructions, and appropriately format the input to engage the tool. In subsequent runs, the executor may receive feedback on previous execution results, then it should adjust the hyperparameters according to these previous results to enhance future outcomes. The entire process can be formulated as follows:
\begin{align}
    o_{j+1} = \mathcal{E} (s_j, o_j, f_j), \quad j=1,2,\ldots, n,
\end{align}
In this process, $o_j$ and $f_j$ represent the previously generated results and feedback, respectively. The system output is defined as $O = o_{n+1}$.

For instance, during the initial run, the Executor may not effectively ``add a hat'' due to the use of inappropriate classifier-free guidance. In response to the feedback signal ``the hat has not been added'', the Executor may enhance the classifier-free guidance to improve performance.

\subsection{Discriminator Agent}\label{sec:discriminator}

Evaluating the results is a crucial step towards their improvement. Hence, we employ a Discriminator Agent which serves a dual purpose. Firstly, it is responsible for assessing the quality of the edited images and providing valuable feedback that contributes to the enhancement of these results. Secondly, it is tasked with selecting the best results to present to the user.

\subsubsection{Generate Feedback}\label{sec:feedback}

Given the caption $\mathcal{C}$ of the input image and the user request $R$, we can design several questions to assess whether the generated image meets the stipulated requirements.  
Questions assess specific request items and overall editing quality in two parts.
The first part verifies if edits match user requirements, typically prompting a binary ``Yes/No'' and explanation.
The second part rates overall quality, considering naturalness and aesthetics.

The agent sequentially answers these evaluative questions.
For questions concerning specific editing items, the agent enlists help from LLMs with visual question-answering capabilities, such as LLaVA \cite{liu2023improvedllava15} or GPT-4V~\cite{openai2023gpt4}. These models take the edited image and question as input and output the answer. As for the comprehensive quality assessment, we not only rely on these types of LLMs, but we also incorporate the Aesthetic Predictor~\cite{aesthetic-predictor} to evaluate the naturalness and visual appeal of the output.

We compile responses into a succinct feedback report.
The feedback offers clear, actionable directions for editing enhancement.
The entire process can be formalized as follows,
\begin{align}
    F = \mathcal{FB} (O, R),
\end{align}
where $\mathcal{FB}(\cdot)$ is the agent to generate feedback $F$, and $O$ denotes the output from the generator.

To enhance the generator's ability to reflect and improve through feedback, we undertake two measures: Firstly, we transmit the feedback from both generator agents back to their respective origins, enabling them to learn from each other's successful strategies or avoid redundant exploration. Secondly, we dissect the overall feedback for each subtask $s_j$. This process enables the planner to more effectively discern the appropriateness of the goal and tools with each subtask. 
Additionally, the specific feedback guides the executor in fine-tuning hyperparameters, even with unchanged goals.

Consider the first generator agent in Fig.~\ref{fig:framework}, in the first round, the feedback is ``A rainbow is visible in the sky, there are no sunflowers in the field, and there is no indication of a wooden barn in the image. Unpleasant visual artifacts are present in the photo''.
The decomposed feedback for each subtask is: 1. Rainbow is successfully added; 2. There are no sunflowers in the field, suggest changing the tool to \texttt{GroundingDINOInpainting}; 3. There is no wooden barn in the image.  

\subsubsection{Quality Competitor}\label{sec:quality-competitor}

In order to achieve results of higher quality and robustness, we leverage two generator agents to generate results and engage a quality Competitor agent $\mathcal{QC}$ to choose the superior one. For each edited image $O$, the competitor agent should compare their corresponding feedback $F$, and select the one that best aligns with the user's request. 

In addition to competing in the generated results of these two agents, competition can also occur across different rounds. Specifically, the agent maintains a memory bank $\mathcal{M}$ to update the current best result $\mathcal{I}_{\text{best}}$. This process can be formalized as:
\begin{align}
    \mathcal{I}_{\text{best}}, F_{\text{best}} &= \mathcal{QC}\left( \left\{ O, F \right\}, \mathcal{M} \right), \\
    \mathcal{M} &= \left\{ \mathcal{M}, \left( \mathcal{I}_{\text{best}}, F_{\text{best}} \right) \right\},
\end{align}
In this formula, $\mathcal{QC}(\cdot)$ represents the quality competitor function that determines the best result. $\left\{ O, F \right\}$ is the set of different generator agents' output and feedback at the current run.

The competitor agent also plays a crucial role in deciding when to terminate the process. In each round, the agent checks if the current best result sufficiently meets the user's requirements. It is not necessary to proceed through all the rounds if the image quality is already deemed satisfactory. As discussed in Sec.~\ref{sec:quality-competitor}, the Quality Competitor plays an important role in obtaining the best result and facilitating early termination. 

\begin{figure*}[!t]
    \centering
    \begin{minipage}[t]{.45\textwidth}
        \centering
        \begin{tcolorbox}[colback=red!5!white,colframe=red!40!white,title=“What” question]
            \begin{small}
                \textit{Question}: What kind of dog is in the image? \\
                \textit{Answer}: There is a dog and a woman sitting on the floor. The dog has white fur. \\
                \textit{Suggestion}: Change the dog to a corgi.
            \end{small}
        \end{tcolorbox}
    \end{minipage}%
    \hfill
    \begin{minipage}[t]{.545\textwidth}
        \centering
        \begin{tcolorbox}[colback=red!5!white,colframe=blue!40!white,title=“Yes/No” question]
            \begin{small}
                \textit{Question}: Does the dog in the edited image look convincingly like a Corgi? \\
                \textit{Answer}: Yes, the dog has been changed to the corgi. \\
                \textit{Suggestion}: Executed successfully, keep the subtask unchanged.
            \end{small}
        \end{tcolorbox}
    \end{minipage}
    \caption{The example user requirements is: ``Change the dog to corgi and transform the image to pixel style''. Yes/No questions can achieve more effective feedback.}
    \label{fig:abl-yes-no-qa}
\end{figure*}

\subsection{Collaborative Competitive Agents}\label{sec:collaboration-competition}

Our whole system contains two generative agents and one discriminator agent. The discriminator agent selects the best result for the user and provides feedback to the generative agents. 
Generative agents refine their strategies using both self-feedback and insights from peers.
For example, if the first agent's plan is to ``Perform subtask A first and then subtask B'', and the second plan is to reverse the order of subtasks. If the discriminator agent recognizes that the second one gets a better result, the first agent may learn from this feedback to change the order for better performance. 
This demonstrates agents' cooperation via shared feedback.
In Sec.~\ref{sec:abl-colla-comp}, we demonstrate that such collaboration will make the system more robust and achieve better results.

In addition to cooperation, the discriminator will also promote competition between the two generative agents. During the initial stage of plan generation, different agents generate various outcomes due to randomness. These variations result in distinct edited results and subsequent feedback. Agents that produce poor results use feedback to improve their results, trying to produce better results than their counterparts.
The discriminator agent benefits from this competitive mechanism as well. By learning from the edited images and feedback generated by various agents, the discriminator agent can provide more refined feedback and suggestions to the generator agent. If there's no discernible difference between the edited results, the discriminator agent will suggest selecting an alternative, more suitable tool to accomplish the sub-goal. The whole algorithm of Collaboration Competition Agent is shown in Alg.~\ref{alg:cca}.

\subsection{Hierarchical Tool Configuration}\label{sec:tool}

It's a huge challenge for agents, especially generator agents, to understand and accurately utilize various tools. Thus we propose a hierarchical tool configuration. For each tool, it should comprise a tool name, a description, and a user manual. The description succinctly articulates the tool's functionality, typically in one or several sentences. 
The manual offers detailed usage instructions, parameter effects, and input/output specifications.

Given the tool diversity, manuals are detailed.
The Planner agent only reads the description of all the tools, decompose user request into sub-goals, and choose an appropriate tool for each sub-goal. The Tool Executor agent takes the user manual of the corresponding tool as input and designs the necessary parameters to use the tool.

\section{Experiments}

In this section, we initially explore the challenges inherent in the construction of such a system, providing an in-depth analysis of several key components in the process. Subsequently, we perform an ablation study to evaluate the efficacy of our individual components. Lastly, we compare our method with other image editing techniques to highlight the advantages of our approach.  

\subsection{Implementation Details}\label{sec:tool-detail}

\begin{figure*}[!t]
    \centering
    \begin{minipage}[t]{.4\textwidth}
        \centering
        \begin{tcolorbox}[colback=red!5!white,colframe=red!40!white,title=w/o Hierarchical tool setting]
            \begin{small}
                \textit{Plan}: 1. Use \texttt{InstructDiffusion} to add wooden frames to the photo, text classifier-free guidance is 2.0; \\
                2. \texttt{Resize} the longer side of the image to 512.\\
            \end{small}
        \end{tcolorbox}
    \end{minipage}%
    \hfill
    \begin{minipage}[t]{.595\textwidth}
        \centering
        \begin{tcolorbox}[colback=red!5!white,colframe=blue!40!white,title=Hierarchical tool setting]
            \begin{small}
                \textit{Plan}: 1. Use \texttt{ImageExpand} to add a 50-pixel white border on all sides of the image; \\
                2. Use \texttt{SDXL-Inpainting} to inpaint this area with wooden frames, with classifier-free guidance 4.0, inpainting prompt ``photo surrounded by wooden frames''; \\
                3. \texttt{Resize} the longer side of the image to 512.
            \end{small}
        \end{tcolorbox}
    \end{minipage}
    \caption{The example of the effect of the hierarchical tool setting. The given user request is ``Enrich wooden frames to the photo and adjust the longer side to 512''.}
    \label{fig:abl-example-tool}
\end{figure*}

We aim to build an automatic system to complete task A. The two most important parts are what kind of ``brain" is used to think about the problem, and what tools are used to complete the task. For the Planner and Feedback part, we leverage GPT-4 to develop plans, generate feedback and suggestions. For the Tool Executor and Quality Competitor, we adopt GPT-3.5-turbo for its speed.

The type and quality of the toolset directly determine the complexity of tasks that can be accomplished and the quality of their completion. We furnish our generator agents with a diverse set of 20 tools, which fall broadly into several categories: Image Preprocessing, Localization, Understanding, Conditional Generation, and General Editing. For the discriminator agent, we deploy the state-of-the-art, open-source, large multi-modal model (LMM), LLaVA-1.5 \cite{liu2023improvedllava15}, which is designed to understand and evaluate the quality of the edited images. 
LLaVA-1.5 excels in detailed captioning and VQA.
Additionally, we also utilize an Aesthetic Predictor~\cite{aesthetic-predictor} to evaluate the overall perceptual quality of the results. Further details are provided in the supplementary material.

\begin{figure*}[t]
    \centering
    \includegraphics[width=1\textwidth]{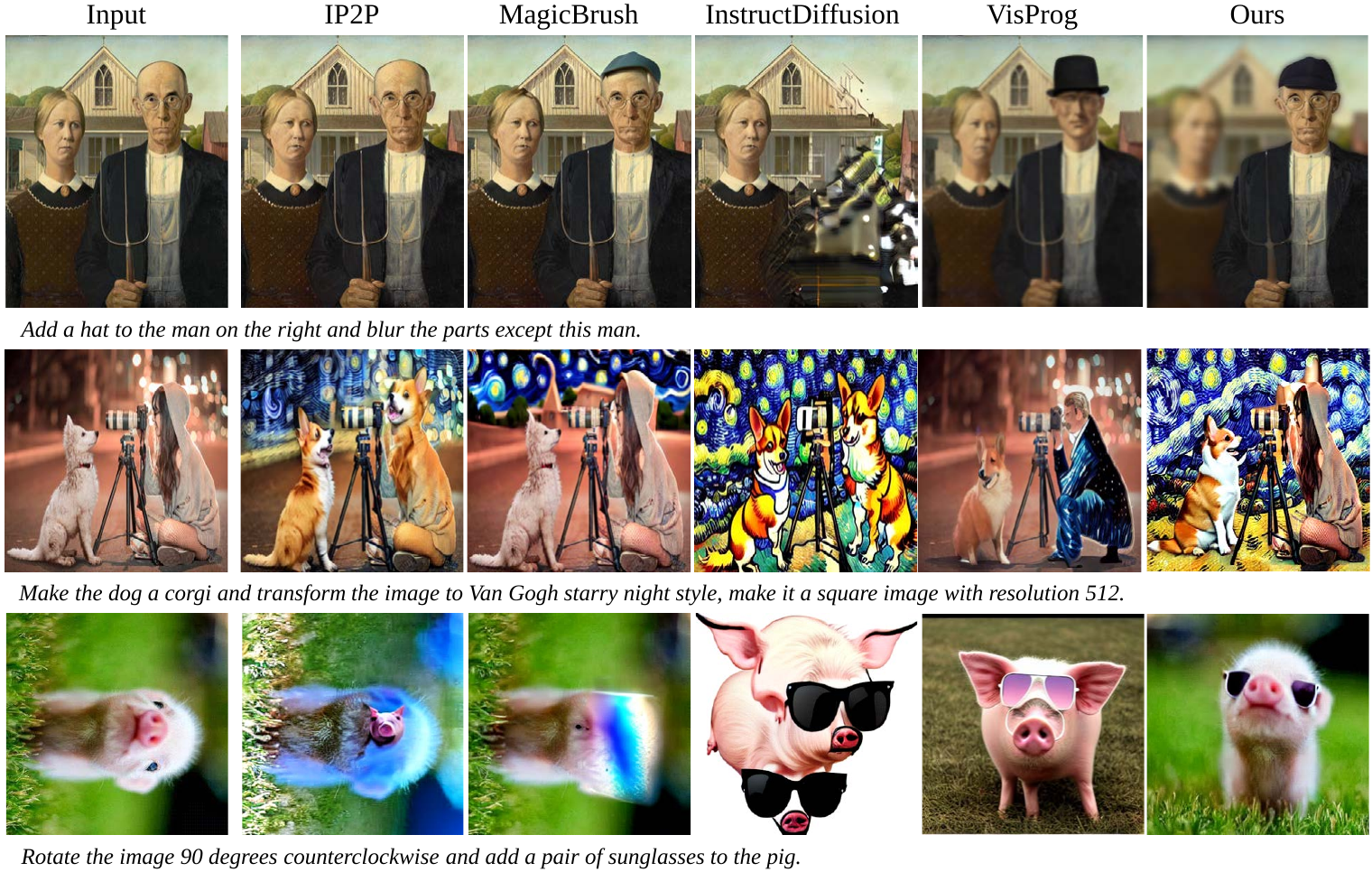}
    \caption{Qualitative comparison between InstructPix2Pix~\cite{brooks2023instructpix2pix}, MagicBrush~\cite{Zhang2023MagicBrush}, InstructDiffusion~\cite{Geng23instructdiff}, VisProg~\cite{gupta2023visual}, and ours.}
    \label{fig:cca-compare}
\end{figure*}

\subsection{Step-by-Step to Build the Framework}

\subsubsection{Yes/No questions rather than what}

Feedback plays a pivotal role in enhancing the quality of editing. To glean information from the edited image, we design several questions and employ LLaVA \cite{wallace2023edict} to answer these questions based on the edited image. 
Question design is key to eliciting precise feedback.
  
Initially, we attempted to ask questions like ``What about the results generated in the figure?''. However, we found that it was challenging for LLaVA to assess the extent of the edited item, and it was also difficult to generate a suggestion according to the vague answer. 
Fig.~\ref{fig:abl-yes-no-qa} showcases that ambiguous questions may lead to confusion about whether the object has been successfully edited. In contrast, ``Yes/No'' questions tend to be answered with greater accuracy.  
Consequently, we modified our approach to pose "Yes/No" questions according to the user requirements, which yielded more precise feedback and better suggestions.

\subsubsection{Tool Diversity}

While several studies~\cite{Geng23instructdiff,brooks2023instructpix2pix} claim the ability to manage diverse types of editing tasks, such as object addition, removal, and replacement, these capabilities fall short of addressing the varied needs of practical applications. Conversely, even when multiple tools are employed for the same task, each may possess its own unique strength, potentially leading to synergistic enhancements. To illustrate this, we choose InstructDiffusion~\cite{Geng23instructdiff} as our baseline method, which can handle a wide range of image editing tasks following user instructions through the diffusion process. Meanwhile, we also incorporate EDICT~\cite{wallace2023edict} and GroundingDINO+Inpainting \cite{liu2023groundingdino,podell2023sdxl} to expand our toolset.  
\begin{figure}
    \centering
    \includegraphics[width=0.5\textwidth]{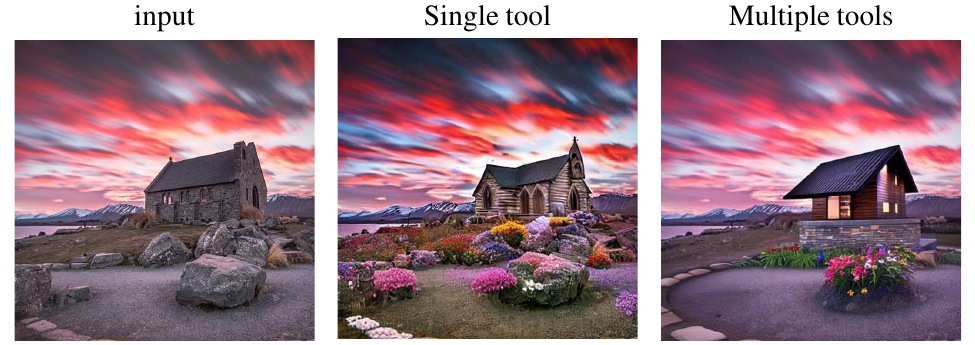}
    \caption{Comparison of single tool vs. multiple tools. Prompt: Replace the house with a wooden one, and turn the stones in front of the house into flowers.}
    \label{fig:single-tool}
\end{figure}
As depicted in Fig.~\ref{fig:single-tool}, the exclusive reliance on a single tool yields subpar result. In this example, the objective is to transform the house into a wooden one and replace the front stone with flowers. 
The single tool with InstructDiffusion failed to locate the front stone while with GroundingDINO's detection ability, multi-tool setting achieved a more reasonable result.

\subsubsection{Stopping Criteria}

In most practical scenarios, we have observed that setting the maximum round, $M$, to 5 is sufficient to meet user requirements. 
Yet, task complexity can greatly change the needed round count.
From a resource conservation standpoint, it is crucial for the quality competitor to determine the appropriate moment to terminate the process. 
We benchmarked this with 20 user requirements, tracking the tool calls by generator agents.

We observed that without the implementation of stopping criteria, the generator agent necessitated an average of 20 tool calls over 5 rounds. However, when we employed the quality competitor's judgment to determine early stoppage, the average number of tool calls reduced to 12 over approximately 3 rounds. Interestingly, we also found that the additional rounds and tool calls did not significantly enhance performance. This might be due to the feedback at this stage not providing explicit guidance on improving the results.

\subsection{Ablation Study}

\subsubsection{Coarse-to-fine Tool Usage}

Given that current large language models such as GPT-3.5-turbo/GPT-4~\cite{ouyang2022instructgpt,openai2023gpt4} possess a finite context length, it presents a challenge for the planner to directly select proper parameters for calling tools while formulating the plan. 
We tackle this with a coarse-to-fine tool usage strategy.
The planner inputs tool descriptions and selects tools, while the executor interprets their specific instructions.
We find that a one-step plan carries an over $20\%$ risk of producing an incorrect format. 
However, this risk diminishes to less than $10\%$ when employing the hierarchical setting.

Taking the user request, ``Enrich wooden frames to the photo and adjust the longer side to 512'' as an example, 
we compare different tool usages in Fig.~\ref{fig:abl-example-tool}. 
Analysis reveals both plans appear fitting, but InstructDiffusion in the first can't manage size changes directly.
In comparison, the hierarchical tool usage breaks the task into subtasks for a logical plan.
This highlights the first approach's inadequate understanding of tool usage.

\begin{figure}
\centering  
\begin{tikzpicture}  
\begin{axis}[  
    width=0.48\textwidth, %
    height=0.3\textwidth, %
    xlabel={Round index},  
    ylabel={txt-cfg},  
    legend style={at={(0.02,0.98)},anchor=north west},  
    xlabel style={yshift=0.5ex},  
    ylabel style={yshift=-3ex},  
    xtick=data,  
    xticklabels={1, 2, 3, 4, 5}  
]  
\addplot coordinates {(1,4.0) (2,5.0) (3,6.0) (4,8.0) (5,8.0)};  
\addplot coordinates {(1,4.0) (2,6.0) (3,7.0) (4,8.0) (5,8.0)};  
\legend{{\footnotesize Agent1}, {\footnotesize Agent2}}
\end{axis}  
\end{tikzpicture}  
\caption{The evolution of classifier-free guidance as the round index increases.}  
\label{fig:abl-instructdiffusion-cfg}  
\end{figure}

Furthermore, we carried out an experiment to verify whether Tool Executor can adjust parameters under the hierarchical tool usage. For a clearer observation, the toolset is restricted to a single tool, \texttt{InstructDiffusion}. We observed that as the number of rounds increases, the Tool Executor gradually increases one of the key parameters $\text{txt-cfg}$ (text classifier-free guidance), as depicted in Fig.~\ref{fig:abl-instructdiffusion-cfg}. It's apparent that the initial value ($\text{txt-cfg}=4$) does not fulfill the requirements and it progressively grows to $\text{txt-cfg}=8$. This observation further underscores the pivotal role that feedback can play in hierarchical tool usage.

\subsubsection{Effect of Collaboration and Competition}\label{sec:abl-colla-comp}

\begin{figure}[!h]
\centering  
\includegraphics[width=0.45\textwidth]{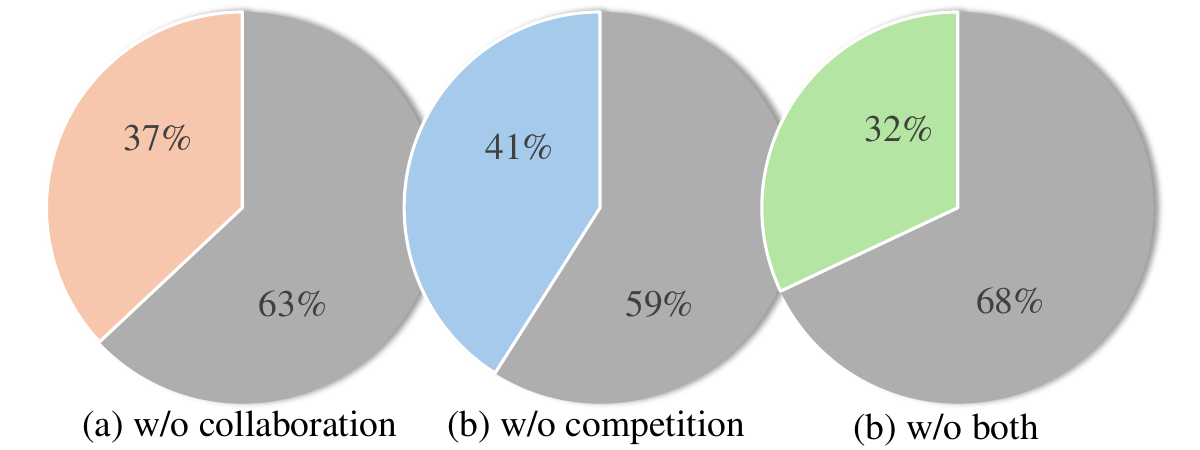}
\caption{Ablation studies of our method by removing collaboration and competition individually, and both together. The grey segment indicates the proportion of users favoring our approach.}  
\label{fig:abl-colla-comp}  
\end{figure}

Our system relies on collaboration for efficient, high-quality planning, and competition to enhance generated results and boost system robustness. 
e tested this across four settings in 100 instances.
a) our full model devoid of both collaboration and competition, b) our model excluding collaboration, c) our model excluding competition, and d) our full model. We compare the first three settings against the final setting, prompting users to determine which results they perceived as superior. The results are demonstrated in Fig.~\ref{fig:abl-colla-comp}. Both strategies contribute to improved outcomes, with the best result achieved by incorporating both elements.

\subsection{Comparison}

We benchmark our CCA against techniques like InstructPix2Pix~\cite{brooks2023instructpix2pix}, MagicBrush~\cite{Zhang2023MagicBrush}, InstructDiffuion~\cite{Geng23instructdiff}, and VisProg~\cite{gupta2023visual}. 
The comparison results, as depicted in Fig.~\ref{fig:cca-compare}, indicate that all previous works struggle to manage such complex cases. Despite VisProg's ability to dissect intricate tasks, it fails to execute edits in alignment with user requirements. 
This underscores the value of using agents cooperatively and competitively.
Additionally, we also perform a quantitative comparison based on human preferences, with the results included in the supplementary material.

\subsection{Discussion}

Our goal is to propose a universal agent-based framework to address complex user requirements rather than focusing on a particular task. Image editing is our first significant achievement on the path toward this goal.  
The CCA framework can also be applied to \textit{other tasks} such as text-to-image generation. We leave this application in the supplementary material. Within our proposed CCA framework, GPT-4 plays a significant role in planning and reflection. However, our method also performs well when utilizing GPT-3.5 Turbo as an alternative. It demonstrate that our method is robust to the choice of LLMs. Related results will also be included in the supplementary material.

\section{Conclusion}

This paper introduces a novel generative framework, Collaborative Competitive Agents (CCA), that employs multiple LLM-based agents to tackle intricate image editing challenges in practice. The key strength of the CCA system lies in its capacity to decompose complex tasks using multiple agents, resulting in a transparent generation process. This enables agents to learn from each other, fostering collaboration and competition to fulfill user requirements. The study's primary contributions entail the proposition of a new multi-agent-based generative model, an examination of the relationships between multiple agents, and extensive experimentation in the area of image editing. This work represents a step forward in AI research, with the potential to influence various fields.

\section*{Appendixes}
\subsection*{Comparison with Previous Methods}

In this section, we conduct a comprehensive comparison of our approach with other methods, encompassing both quantitative results and qualitative results.

\subsubsection*{Human Preference Study}\label{sec:human-preference}

We showcase a human preference study that compares our method with previous state-of-the-art techniques, as demonstrated in Table~\ref{tab:human-prefer}. VisProg~\cite{gupta2023visual} is not included in this table because it is specific design for straightforward object removal tasks and certain low-level operations, typically falls short when subjected to a broader comparison.

We have constructed an interface using Gradio to facilitate this comparison of user preferences. The interface presents the input image, editing instruction, and outcomes from various methods to the user. In addition, we ask human participants to select both the result that best adheres to the instruction and the one that exhibits the highest quality, respectively. The table reveals that 47 percent of respondents concur that our method, CCA, more effectively fulfills user requirements, while 44 percent believe that our results showcase superior quality.

\begin{table}[h]  
\centering  
\begin{tabular}{ccccc}  
\hline  
{} & IP2P & MB & ID & CCA \\ \hline  
Txt-alignment & $11\%$ & $21\%$ & $21\%$ & $47\%$ \\ 
Visual Quality & $16\%$ & $17\%$ & $23\%$ & $44\%$ \\ \hline  
\end{tabular}  
\caption{Human preference. In this table, IP2P, MB, and ID are abbreviations for InstructPix2Pix~\cite{brooks2023instructpix2pix}, MagicBrush~\cite{Zhang2023MagicBrush}, and InstructDiffusion~\cite{Geng23instructdiff}, respectively.  
}  
\label{tab:human-prefer}  
\end{table}  

\subsubsection*{Effect of Number of Agents}

We add some quantitative results in Table~\ref{tab:comparison-num-agents}, and we will add related results in revised version. For CLIP score, system with 2 agents performs better than 1 agent by 1.0 and 3 agents outperforms 2 agents by 0.04. For aesthetic score, 2-agent system performs better than its counterparts. L1 distance measures the distance between input image and edited image. 2-agent system achieves good CLIP score and aesthetic score, and also maintains the distance between input and edited image.

\begin{table}[h]
\centering
\begin{tabular}{@{}lr@{}}
\toprule
\multicolumn{2}{@{}l}{CLIP Score} \\
\midrule
InstructDiffusion & 31.21 \\
1 agents & 31.09 \\
2 agents & 32.09 \\
3 agents & 32.13 \\
\midrule
\multicolumn{2}{@{}l}{Aesthetic Score} \\
\midrule
InstructDiffusion & 6.16 \\
1 agents & 6.20 \\
2 agents & 6.31 \\
3 agents & 6.07 \\
\midrule
\multicolumn{2}{@{}l}{L1 Distance} \\
\midrule
InstructDiffusion & 134.25 \\
1 agents & 130.42 \\
2 agents & 131.22 \\
3 agents & 138.66 \\
\bottomrule
\end{tabular}
\caption{Comparison of different methods}
\label{tab:comparison-num-agents}
\end{table}

\subsubsection*{More Qualitative Results}

\begin{figure*}[p]
    \centering  
    {
    \normalsize
    \begin{tabularx}{\textwidth}{*{5}{Y}}  
        Input & InstructPix2Pix & InstructDiffusion & MagicBrush & CCA 
    \end{tabularx}  
    }
        \includegraphics[width=\textwidth]{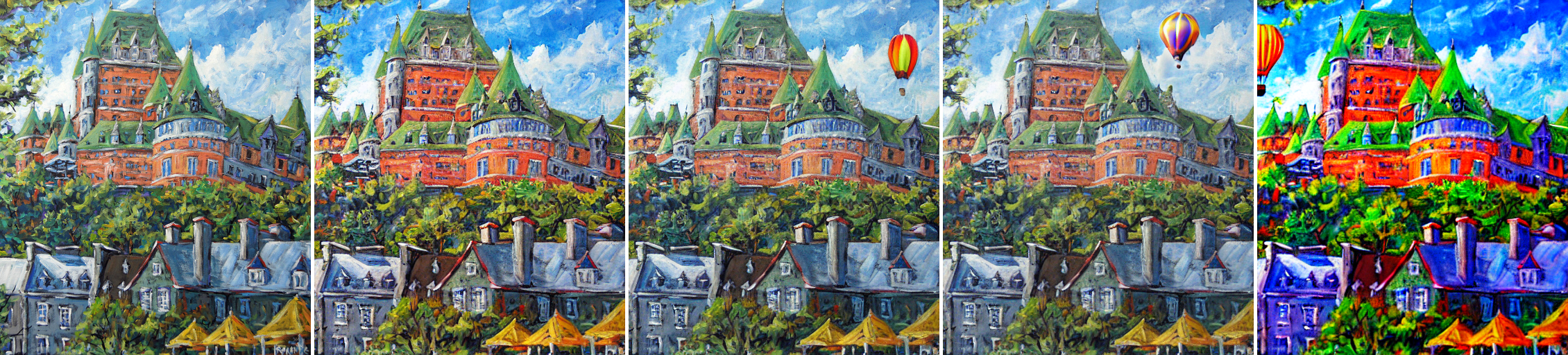}  
        \caption*{{\normalsize Add a hot air balloon in the sky and make the colors more vibrant. }}  
    \hfill   
        \includegraphics[width=\textwidth]{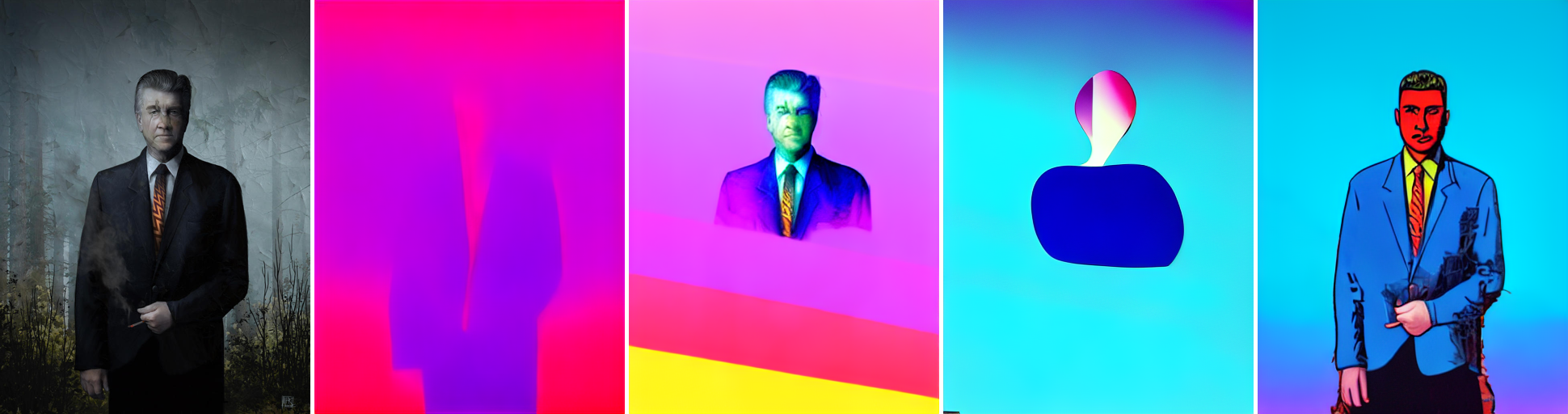}  
        \caption*{{\normalsize Transform the image into a pop art style and replace the background with a vibrant color gradient.}}  
    \hfill  
        \includegraphics[width=\textwidth]{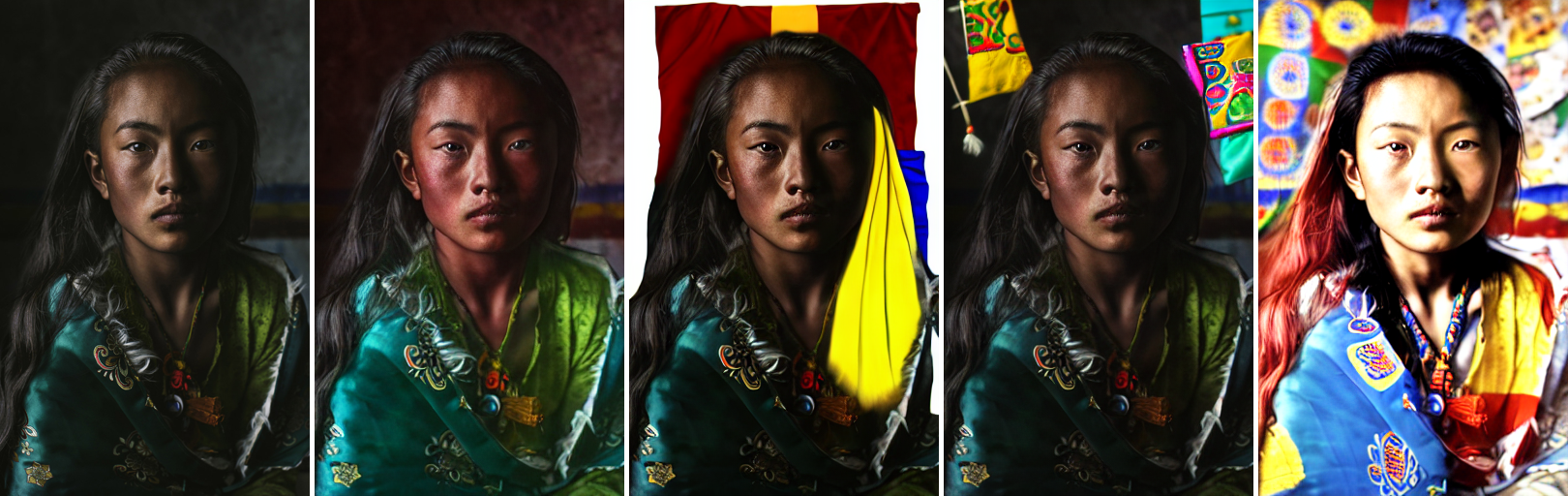}  
        \caption*{{\normalsize Add a traditional Tibetan prayer flag to the background and adjust the color to be more vibrant.}}  
    \hfill  
        \includegraphics[width=\textwidth]{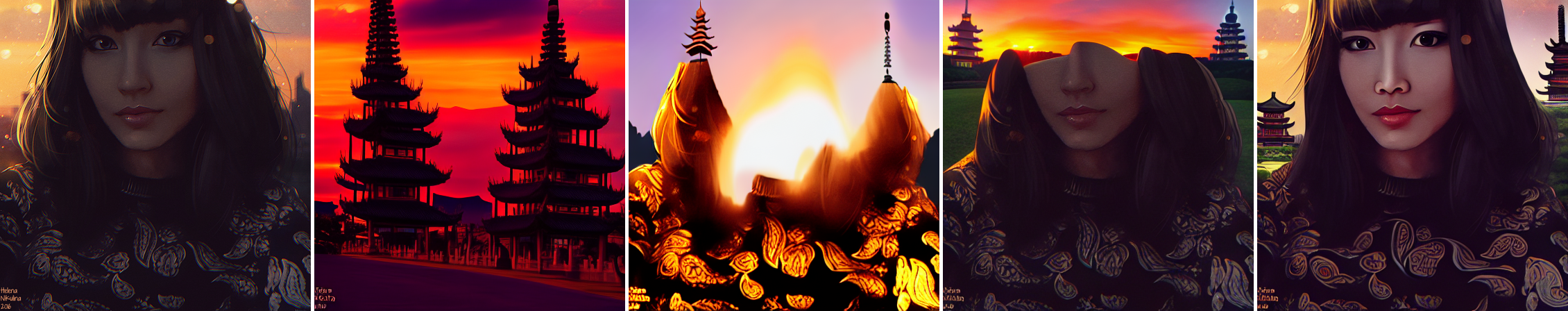}  
        \caption*{{\normalsize Please add a traditional Chinese pagoda in the background and make the colors more vibrant.}}  
\caption{Comparison results between previous methods and our CCA. From left to right: input image, results from InstructPix2Pix~\cite{brooks2023instructpix2pix}, MagicBrush~\cite{Zhang2023MagicBrush}, InstructDiffusion~\cite{Geng23instructdiff}, and our proposed CCA. Below the images is the editing instruction.}  
\label{fig:comp-1}
\end{figure*}

\begin{figure*}[p]
    \centering  
    {
    \normalsize
    \begin{tabularx}{\textwidth}{*{5}{Y}}  
        Input & InstructPix2Pix & InstructDiffusion & MagicBrush & CCA  
    \end{tabularx}  
    }
        \includegraphics[width=\textwidth]{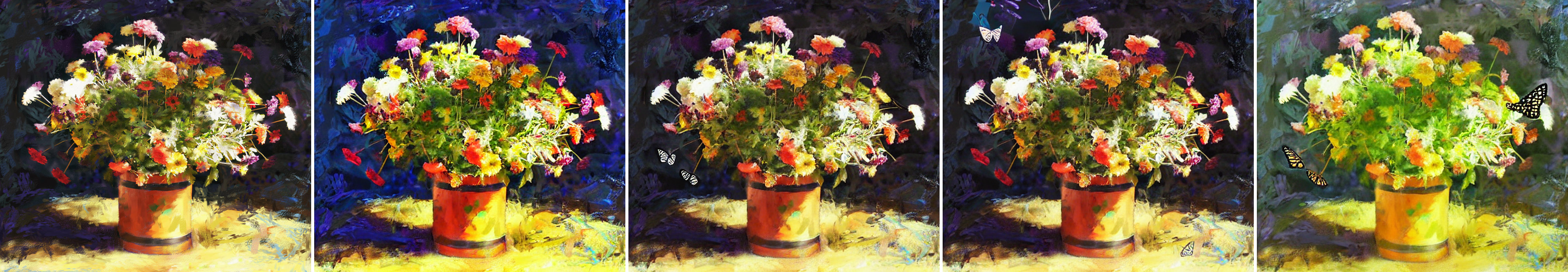}  
        \caption*{{\small Add butterflies in the foreground and make the colors more vibrant. }}  
    \hfill  
        \includegraphics[width=\textwidth]{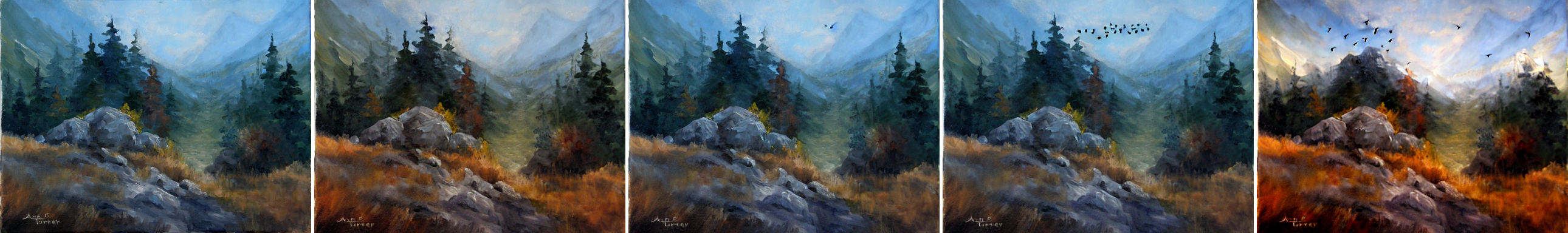}  
        \caption*{{\small Add a flock of birds flying over the mountain pass and change the color palette to warm autumn tones.}}  
    \hfill  
        \includegraphics[width=\textwidth]{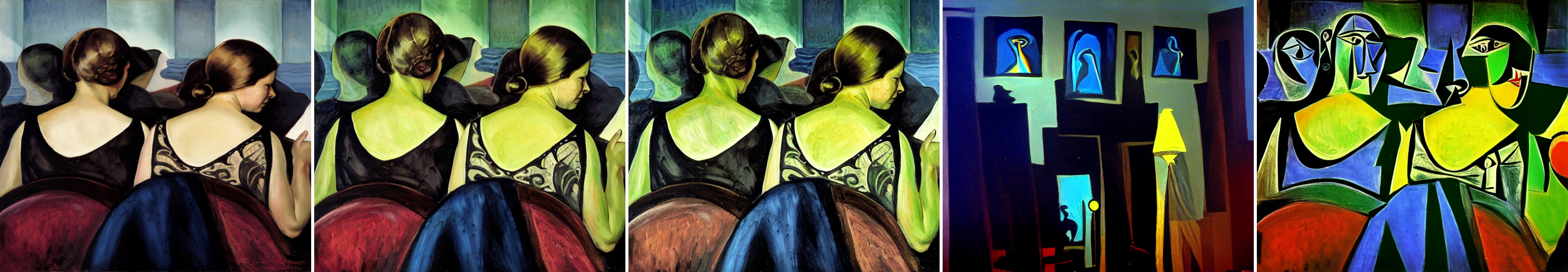}  
        \caption*{{\small Change the image to a night scene, add a spotlight on the main character, and transform the style to be reminiscent of a Picasso painting. }}  
    \hfill  
        \includegraphics[width=\textwidth]{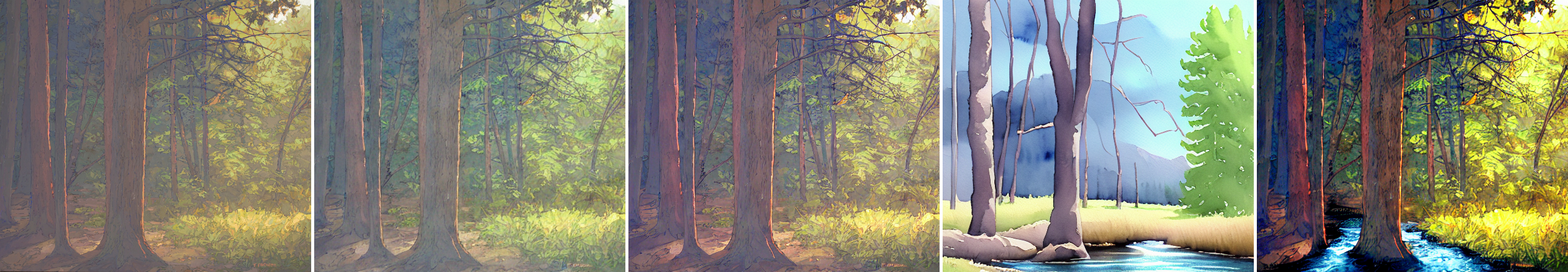}  
        \caption*{{\small Add a small stream flowing through the trees and transform the style to be a watercolor painting.}}  
    \hfill  
        \includegraphics[width=\textwidth]{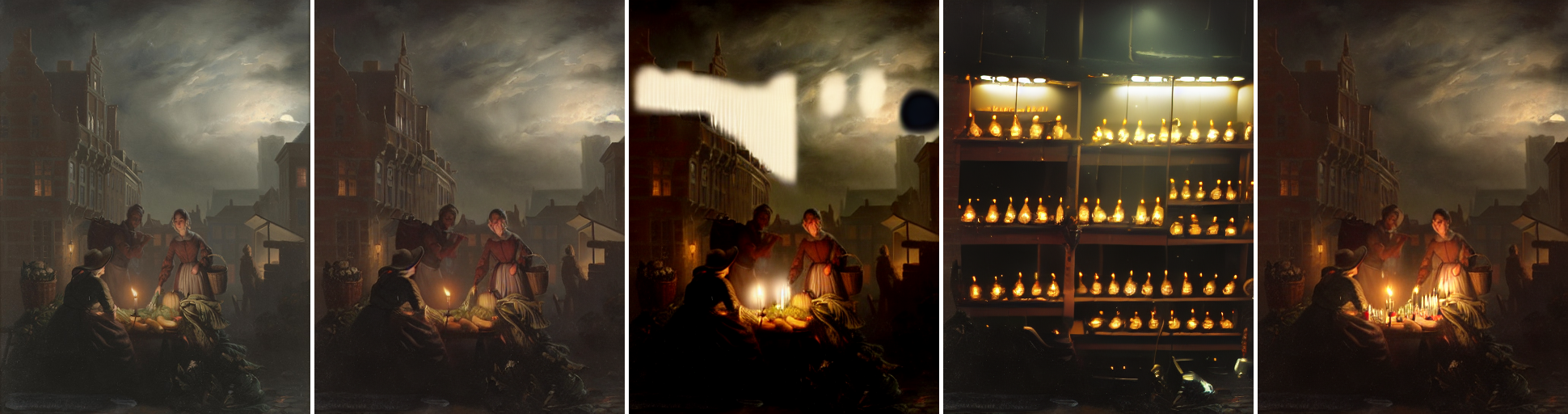}  
        \caption*{{\small Add more candles to the market stall and make the overall lighting warmer and more inviting.}}  
\caption{Comparison results between previous methods and our CCA. From left to right: input image, results from InstructPix2Pix~\cite{brooks2023instructpix2pix}, MagicBrush~\cite{Zhang2023MagicBrush}, InstructDiffusion~\cite{Geng23instructdiff}, and our proposed CCA. Below the images is the editing instruction.} 
\label{fig:comp-2}
\end{figure*}

\begin{figure*}[p]
    \centering
    \centering  
    {
    \normalsize
    \begin{tabularx}{\textwidth}{*{5}{Y}}  
        Input & InstructPix2Pix & InstructDiffusion  & MagicBrush & CCA  
    \end{tabularx}  
    }
        \includegraphics[width=\textwidth]{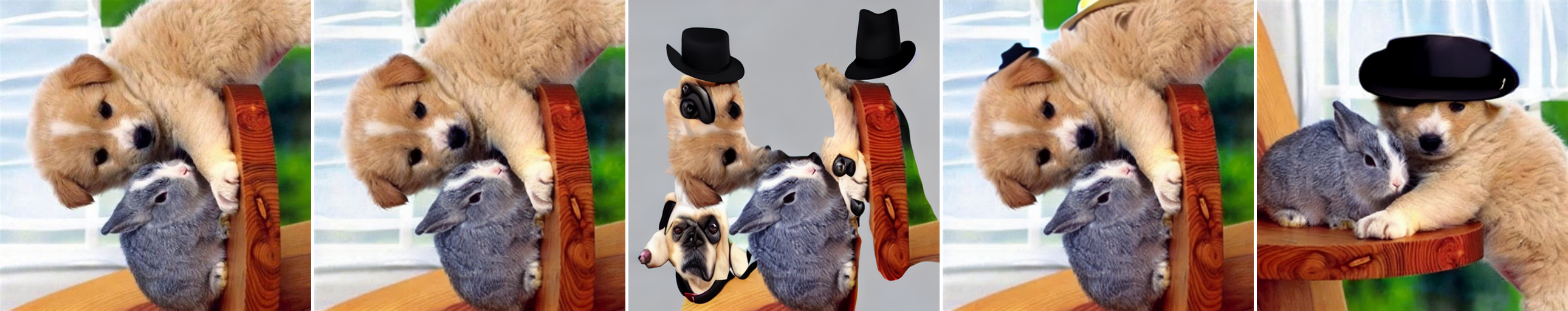}  
        \caption*{{\small Rotate image clockwise, add a hat to the dog }}  
        \includegraphics[width=\textwidth]{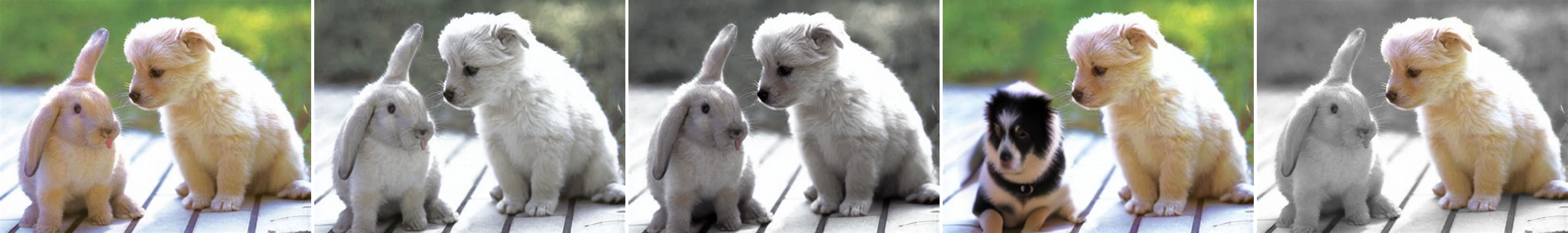}  
        \caption*{{\small Make the image gray except the dog}}  
    \hfill  
        \includegraphics[width=\textwidth]{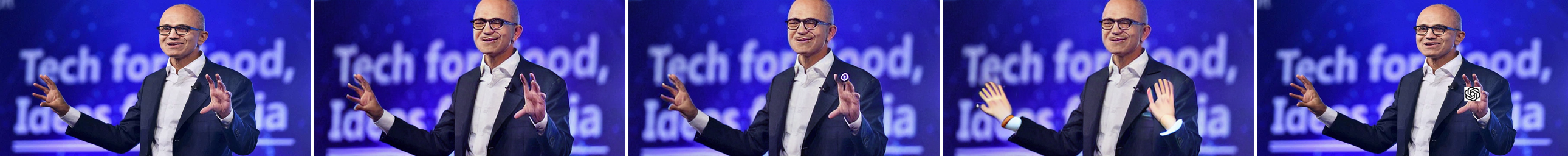}  
        \caption*{{\small Add OpenAI's logo on Satya's one hand}}  
    \hfill  
        \includegraphics[width=\textwidth]{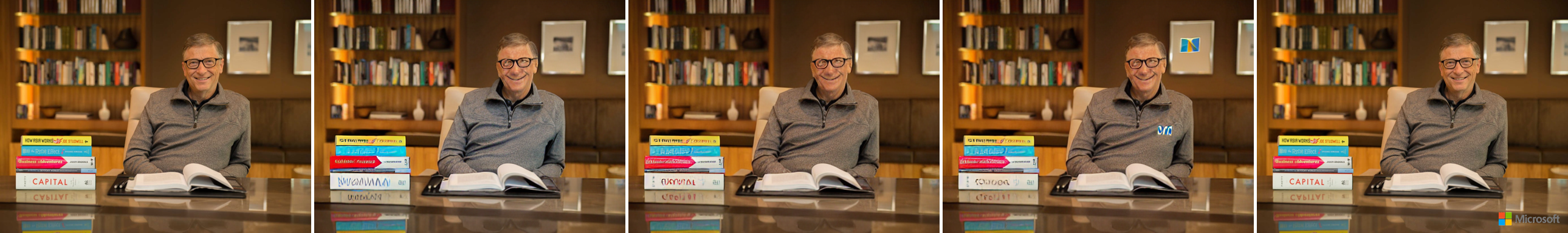}  
        \caption*{{\small Add Microsoft logo as watermark to the image}}  
    \hfill  
        \includegraphics[width=\textwidth]{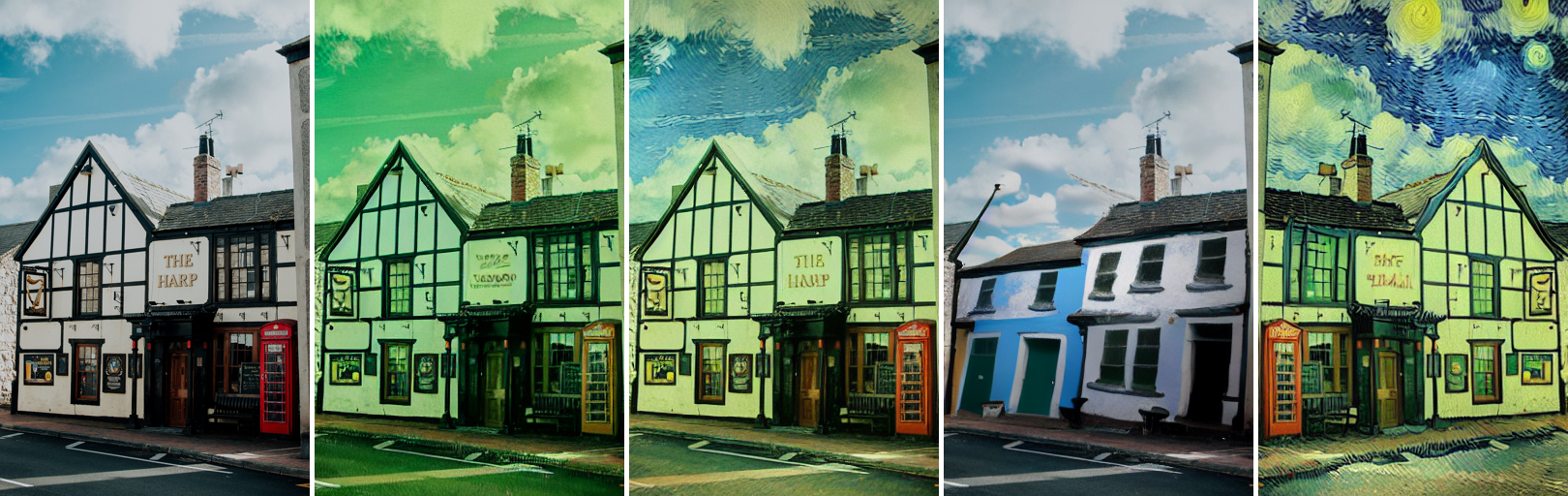}  
        \caption*{{\small Transform the style of the image to be Van Gogh Cottages, then flip it.}}  
    \caption{Comparison results between previous methods and our CCA. From left to right: input image, results from InstructPix2Pix~\cite{brooks2023instructpix2pix}, MagicBrush~\cite{Zhang2023MagicBrush}, InstructDiffusion~\cite{Geng23instructdiff}, and our proposed CCA. Below the images is the editing instruction.
    }
    \label{fig:comp-3}
\end{figure*}

We show more qualitative results in Figure~\ref{fig:comp-1}, \ref{fig:comp-2}, and \ref{fig:comp-3}.
From the first two figures, we compare with previous methods on more cases.
The third figure aims to further underscore the efficacy of our approach. Contrary to conventional image editing techniques, our agent-based system exhibits superior performance in tasks such as background removal, precise black and white conversion of specific regions, and sticker addition.

\newpage
\subsection*{Prompt Templates}\label{supp:prompt}
The prompt supplied to the LLMs can substantially influence its performance. Hence, We introduce the main prompt templates employed within our framework. The necessary input variables are highlighted in italics and enclosed by \textless \textgreater. Further prompts, including tool descriptions and the user manual, will be made available concurrent with the code release soon.

\begin{myshadowbox}
\small
{\color{blue} Planner agent $\mathcal{P}$ (first round of planning)} 

I want to edit the image \textless {\textit{IMAGE PATH}} \textgreater using user's instruction `\textless {\textit{EDITING REQUEST}} \textgreater`. Please help me decompose this task into several subtasks. All images should be saved at the same folder `image/`.\\
            
Each subtask should be short and specific that can be done by only a single tool. Each subtask should only be tried once. Each subtask should be described in a single line. \\
The tool should be one of the following: \\
``` \\
    \textless {\textit{TOOL NAMES}} \textgreater \\
``` \\

The detailed description of each tool is as follows: \\
``` \\
    \textless {\textit{TOOL DESCRIPTIONS}} \textgreater \\
``` \\
            
For example, if the user requirement is `Create a vintage-style portrait of a person with a hat and adjust the image to have a sepia tone, with the longest side being 800 pixels.` with an given input image, it can be decomposed into the following steps: \\
{1. Resize the image to have its longest side at 800 pixels using $\texttt{Resize}$; 2. Add a vintage-style hat to the person in the image using $\texttt{Instructdiffusion}$; 3. Apply a sepia tone filter to the entire image $\texttt{Edict}$.} \\

Do not include specific input/output image path in subtasks. If you have to resize the image, put it at the first. The final response should be concise and clear. \\

Now give me the plan.
\end{myshadowbox}  

\begin{myshadowbox}
\small
{\color{blue} Planner agent $\mathcal{P}$ (reflection)}

    I want to edit the image \textless {\textit{IMAGE PATH}} \textgreater using `\textless {\textit{{EDITING REQUEST}}} \textgreater`. Please help me decompose this task into several subtasks. All images should be saved at the same folder `image/`\\
            
        Each subtask should be short and specific that can be done by only a single tool. Each subtask should only be tried once. Each subtask should be described in a single line.\\

        The tool must be one of the following:\\
        ```\\
        \textless {\textit{TOOL NAMES}} \textgreater\\
        ```\\

        The detailed description of each tool is as follows:\\
        ```\\
        \textless {\textit{{TOOL DESCRIPTIONS}}} \textgreater\\
        ```\\

        Currently, I have decomposed it into the following plans {(subtask with related tool)}:\\
        ```\\
        \textless {\textit{{SUBTASKS}}} \textgreater\\
        ```\\

        The feedback obtained by executing step by step is:\\
        ```\\
        \textless {\textit{{FEEDBACK}}} \textgreater\\
        ```\\
        Besides, we also have another plan:\\
        ``` \\
        \textless {\textit{PLAN}} \textgreater\\
        ```\\
        This plan obtained the following feedback: \\
        ```\\
        \textless {\textit{FEEDBACK}} \textgreater\\
        ```\\
        Comparing the current plan and the referenced plan, Do you think I should change the order of the subtasks or modify the content of subtasks? If yes, please tell me the new plan to improve the editing quality. You should choose one specific tool for each subtask. If you think I should only change the tool or the input to the tool, please respond with ``No''.
\end{myshadowbox}

\begin{myshadowbox}
\small
    {\color{blue} Feedback agent $\mathcal{FB}$ (question generation)}

    We want to edit the image using instruction `\textless {\textit{EDITING REQUEST}} \textgreater`. The detailed description of the input image is `\textless {\textit{CAPTION}} \textgreater`.
        
Suppose we have edited the image, please design some questions to ask human to judge the quality of the edited image. \\

For example, if the task is `Transform a daytime cityscape photo into a nighttime scene with lit streetlights and a full moon, with the long side being 800 pixels.`. The description of the image is `A cityscape photo with a busy street, tall buildings, and people walking around.` Then the questions can be like: 1. Is the overall setting of the picture a nighttime scene? \textbackslash n  2.Are streetlights visible in the original image?\textbackslash n  3.Is the moon present in the photo? \textbackslash n 4. Is the size of image changed to 800?  \\

The questions should be concrete and clear and do not include hallucination. **Yes or No questions are preferred**. Reduce overlap between the questions.  The questions should be composed of local and global editing effects. You need to ensure that parts unrelated to the editing requirements remain unchanged.\textbackslash n The number of questions should be no more than five. Each question should be  concise and clear.\\

Now give me the questions.
\end{myshadowbox}

\subsection*{Tool Set}\label{supp:toolset}
Agents employ tools to accomplish specific tasks. Our goal is to build an automated pipeline for image editing where tools serve as critical components. For customized editing tasks, we first preprocess the image, followed by the selection of suitable tools. We then configure the associated parameters and ultimately obtain the results. We provide the following tools for the agent:

\begin{itemize}
    \item \textbf{Resize}. This function alters the dimensions of an image to a specified resolution. It accepts the image and the desired resolution as inputs and produces a resized image as the output.  
  
    \item \textbf{Paste}. This feature allows for the insertion of a smaller image onto a specified position within a larger one. It requires the base image, the secondary image, and the position coordinates for pasting as inputs, resulting in a composite image as the output.  
    
    \item \textbf{Blending}. Merge two images at a specific area of the base image. It takes the base image, the image to blend, the position coordinates where blending should start, and a blending strength parameter as input. This function outputs the base image with the second image blended into the specified area.  
    
    \item \textbf{InstructDiffusion}~\cite{Geng23instructdiff}. A text-guided image editing tool that accepts an image and an editing prompt as inputs and generates the edited image as output.  
    
    \item \textbf{LLaVA}~\cite{liu2023improvedllava15}. Give the detailed caption of the image and answer questions about the image. It takes the image and question as input and outputs the answer to the question.
    
    \item \textbf{AestheticScore}~\cite{aesthetic-predictor}. Used to assess the aesthetic quality of the image. It takes the image as input and outputs the aesthetic score.
    
    \item \textbf{ImageDifferenceLLaVA}. Based on LLaVA 1.5~\cite{liu2023improvedllava15}, this tool concat the input image and edited image and tells the difference between two parts. It takes the original image and the edited image as input, and returns the string describes the differences between them.
    
    \item \textbf{GroundingDINO}~\cite{liu2023groundingdino}. Utilizing detection prompt, this tool identifies and segments specific elements within an image. It requires an image and a corresponding detection prompt as inputs and outputs the mask.

    \item \textbf{Prompt2Prompt}~\cite{hertz2022prompt-to-prompt}. Designed for text-guided image editing, this tool adeptly handles tasks such as object replacement by taking an original image, a base prompt, and a target prompt as inputs to produce the edited image.  
    
    \item \textbf{Crop}. This tool is designed to selectively crop a specified region of an image, requiring the image and the target region's coordinates as inputs to deliver the cropped output.  
    
    \item \textbf{RGB2Gray}. This tool converts RGB color images into grayscale, accepting a color image as input and producing a single-channel grayscale image as output.
    
    \item \textbf{GaussianBlur}. This function applies a Gaussian blur to the input image. It requires the image and the Gaussian kernel size to return the blurred image. 
    
    \item \textbf{RotateClockwise}. Rotate the image in the clockwise direction. It takes the image as input and outputs the rotated image.
    
    \item \textbf{RotateCounterClockwise}. Rotate the image in the counterclockwise direction. It takes the image as input and outputs the rotated image.
    
    \item \textbf{EnhanceColor}. It is used to adjust the color balance of an image. It takes the image and an enhancement factor as input and outputs the enhanced image. %
    
    \item \textbf{FlipHorizontal}. This operation mirrors the image along the horizontal axis. It accepts an image as input and outputs the horizontally flipped version.

    \item \textbf{AddLogo}. This utility embeds a logo onto an image at a specified location. It requires the original image, the logo file, and the coordinates for the logo's placement as inputs to output an image with the logo superimposed.  
    
    \item \textbf{AddWatermark}. This function overlays a watermark onto an image, specifically positioning it in the bottom right corner. It takes the original image and a watermark image as inputs, with an optional alpha parameter to adjust the watermark transparency. The watermark is blended over the original image, ensuring visibility while maintaining the integrity of the image content.

    \item \textbf{{Inpainting}}: This tool performs inpainting on a specified masking area using SDXL-Inpainting. It takes the image, mask, and target prompt as inputs, and outputs the image whose masked region is inpainted to the target prompt.
    
    \item \textbf{GetSize}: This tool gets the size of the image. It takes the image as input and outputs the size of the image.
    
\end{itemize}

\subsection*{Application of Other Tasks}

\begin{figure}[ht]
    \centering
    \includegraphics[width=0.49\textwidth]{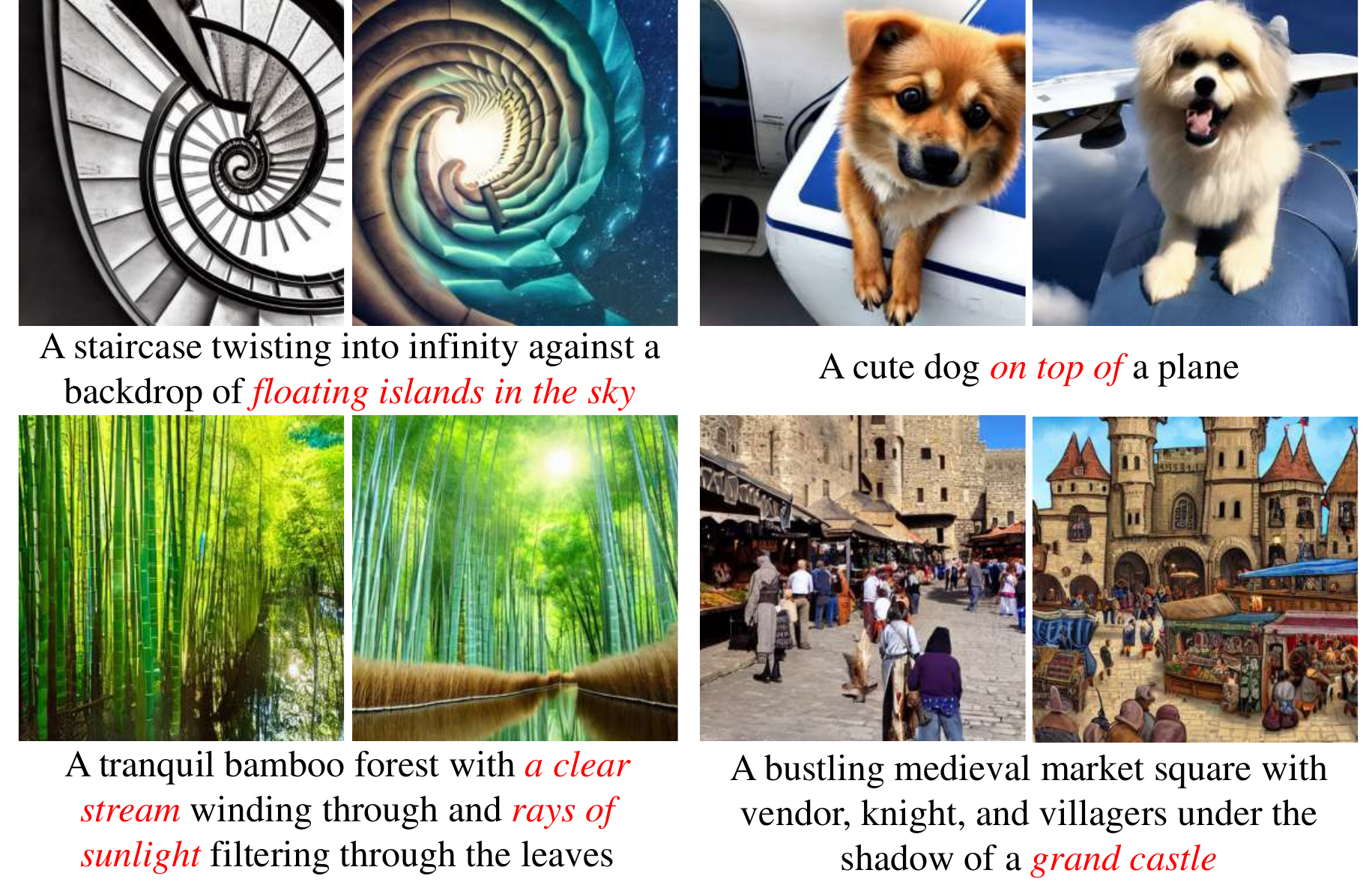}
    \caption{Additional application of CCA on text-to-image task. For each case, the left is generated by Stable Diffusion 1.5 and the right is generated by our framework.}
    \label{fig:t2i}
\end{figure}

We chose the image editing task for our paper because it is relatively easy to obtain feedback from the generated results. Besides, it necessitates the capability to analyze complex instructions for planning, execution, and feedback integration, which highlights the advantages of our framework more effectively. We have also explored the text-to-image task and present some preliminary results in Fig.~\ref{fig:t2i}. When given a prompt, our CCA can produce images that align more closely with the prompt than when a tool is used directly. This advantage is primarily due to the feedback and iterative optimization mechanisms within the CCA system. We will include more details in the final version of this document. These tasks suggest the practicality of our method. We are optimistic about the general applicability of our approach and intend to explore its potential in a broader array of tasks, including robotics and multi-agent systems, in future research.

\vspace{-0.5em}

\subsection*{Running Time Analysis}

We obtained the processing times for different configurations of Large Language Models (LLMs), notably including GPT-3.5-Turbo and GPT-4-Turbo. Twelve experimental trials were conducted to ascertain the duration of each. After discarding the outliers—the most and the least time-consuming trials—we computed the mean processing time. The averaged duration for the GPT-3.5-Turbo amounted to 4.6 minutes, whereas for the GPT-4-Turbo, it requires about 11.5 minutes.

\subsection*{Limitations}\label{supp:limitations}

While the CCA attains commendable results, our experiments reveal two primary limitations. First, the entire pipeline processes the subtasks in a sequential manner. This means that each task relies solely on the output of its preceding subtask, without access to earlier images, which could potentially lead to error accumulation. Second, agents may lack a comprehensive understanding of the tool, necessitating multiple rounds of exploration to achieve satisfactory results. In future work, we plan to introduce a memory module to agents so that they can deepen the understanding of the tool during the reflection process.

\subsection*{Broader Impact}\label{supp:impact}

{
The novel generative model introduced in this paper, the Collaborative Competitive Agents (CCA), leverages multiple Large Language Models (LLMs) to perform complex tasks and demonstrates a significant advancement in image editing. While our work primarily aims to enhance the capabilities of generative models, the broader implications of this research must be carefully considered, including the potential long-term impacts on society, the environment.

Postive Impact:

\begin{itemize}
    \item Innovation in Creative Industries: The CCA model can significantly improve the efficiency and quality of content creation in art, entertainment, and design, leading to new products and services.
    \item Enhanced Decision Making: By decomposing complex tasks, the CCA system can aid in areas requiring intricate analysis, such as climate modeling, urban planning, and logistics.
    Accessibility in Technology: The system's ability to interpret and execute detailed instructions can make sophisticated technological tools more accessible to a broader audience, democratizing design and creativity.
\end{itemize}

Negative Impact:
\begin{itemize}
    \item Misuse of Synthetic Media: The ability of the CCA to create realistic and complex images raises concerns about the production of synthetic media for disinformation, identity theft, or other malicious intents.
    \item Bias Propagation: If the LLMs or datasets used in the CCA inherit societal biases, the generated content might perpetuate or exacerbate these biases, leading to unfair or discriminatory outcomes.
\end{itemize}

The CCA model presents a powerful tool for generative tasks, with the potential to contribute positively to various sectors. However, it is crucial to anticipate the ethical and societal challenges posed by such advancements. It is the responsibility of the research community to engage in ongoing discussions about AI ethics and to develop strategies that mitigate risks and ensure the responsible use of AI technologies.
}

\subsection*{Success rate of various open-source models }

For each planned task, the agent views all tool names and descriptions and selects the appropriate tool. Tool name is extracted from the agent's response.
Given the selected tool name and plan, the agent views the detailed cookbook for the tool and provides the input arguments. For example, if the tool is tool-1, the related arguments are arg1, arg2, and arg3. We require the agent to respond in a specific format, e.g., tool-1 @@ $\text{arg1} \leftrightarrow \text{arg2} \leftrightarrow \text{arg3}$. This structured format facilitates the extraction of arguments and the subsequent execution of the tool.
The hierarchical design enhances the success rate of formatting tool execution.

We evaluate the success rate of various models in the following table. We assessed 100 editing prompts to determine whether the LLM could follow our instructions to generate the formatted plan and subsequently format the appropriate tool usage for each sub-plan (tool-1 @@ $\text{arg1} \leftrightarrow \text{arg2} \leftrightarrow \text{arg3}$).

\begin{table}[h]
\centering
\begin{tabular}{|l|c|}
\hline
Model & Success Rate \\
\hline
Mistral-7B-Instruct-v0.1 & 0.18 \\
Mistral-7B-Instruct-v0.2 & 0.74 \\
Qwen2-7B-Instruct & 0.97 \\
DeepSeek-V2 & 1.0 \\
\hline
\end{tabular}
\caption{Model Success Rate}
\label{tab:model_success_rate}
\end{table}

\subsection*{Additional visual results of various open-source models }

\begin{figure*}
    \centering
    \includegraphics[width=\linewidth]{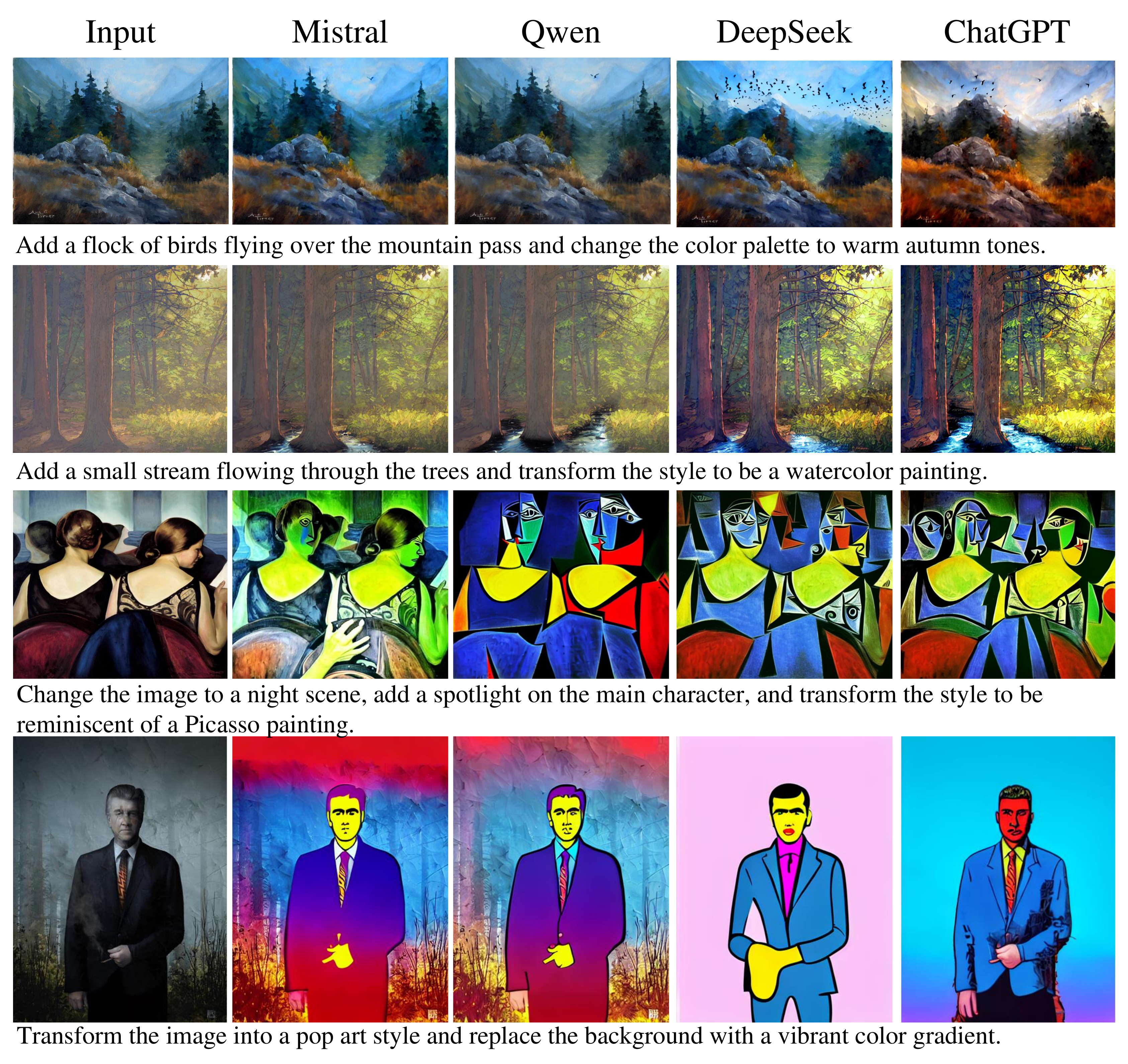}
    \caption{Visual comparison of image editing capabilities across different large language models. From left to right: Input image, Mistral, Qwen, DeepSeek, and ChatGPT results given various editing prompts.}
    \label{fig:results-of-additional-model}
\end{figure*}

We also provided several visual examples to demonstrate the comparative performance of different open-source models on image editing tasks in our framework. As shown in Figure \ref{fig:results-of-additional-model}, we tested various editing scenarios to evaluate the models' capabilities. The results reveal notable differences in performance:
Mistral-7B shows basic editing capabilities but often struggles with complex modifications and maintaining image coherence, particularly in cases requiring sophisticated semantic understanding or fine-grained adjustments.
Qwen2-7B-Instruct demonstrates significantly better performance, with superior handling of complex editing tasks, particularly in maintaining context consistency and generating realistic modifications. The visual results show improved detail preservation and more natural transitions in edited regions.
DeepSeek-V2 produces the most sophisticated editing results among the open-source models. The visual examples highlight its capability to handle nuanced modifications while preserving image quality and semantic coherence, especially in challenging scenarios involving multiple objects or complex transformations.
ChatGPT's results are included as a commercial baseline for comparison, showing comparable performance to DeepSeek-V2 in most cases, though with notably higher computational requirements and associated costs.
These visual comparisons demonstrate the rapid progress in open-source models' capabilities, particularly in image editing applications. The qualitative improvements in newer model versions suggest promising directions for future development in this domain.

\bibliographystyle{fcs}
\bibliography{ref}

\end{document}